\documentclass[11pt,a4paper]{article}
\usepackage[nohyperref]{acl2020}
\usepackage{times}
\usepackage{latexsym}
\usepackage{breakurl}
\usepackage{csquotes}
\usepackage{multirow}
\usepackage{booktabs}
\usepackage{microtype}

\aclfinalcopy 

\title{Synthesizing Monolingual Data for Neural Machine Translation}

\author{Benjamin Marie\qquad Atsushi Fujita  \\
  National Institute of Information and Communications Technology \\
	3-5 Hikaridai, Seika-cho, Soraku-gun, Kyoto, 619-0289, Japan \\
  {\{bmarie, atsushi.fujita\}@nict.go.jp}}

\date{}

\begin{document}
\maketitle
\begin{abstract}
In neural machine translation (NMT), monolingual data in the target language are usually exploited through a method so-called ``back-translation'' to synthesize additional training parallel data. The synthetic data have been shown helpful to train better NMT, especially for low-resource language pairs and domains. Nonetheless, large monolingual data in the target domains or languages are not always available to generate large synthetic parallel data. In this work, we propose a new method to generate large synthetic parallel data leveraging very small monolingual data in a specific domain. We fine-tune a pre-trained GPT-2 model on such small in-domain monolingual data and use the resulting model to generate a large amount of synthetic in-domain monolingual data. Then, we perform back-translation, or forward translation, to generate synthetic in-domain parallel data. Our preliminary experiments on three language pairs and five domains show the effectiveness of our method to generate fully synthetic but useful in-domain parallel data for improving NMT in all configurations. We also show promising results in extreme adaptation for personalized NMT.
\end{abstract}

\section{Introduction}
\label{section:intro}
Neural machine translation (NMT) systems usually require a large quantity of parallel data for training. For most language pairs and domains, we do not have such resources, or only in very small quantities, mainly because they are costly to produce \citep{W01-1409}.
Unlike parallel data, monolingual data are readily available in large quantity for many languages. Previous work has proposed various strategies to integrate monolingual data into NMT systems and has confirmed their usefulness to improve NMT systems, especially in low-resource configurations. The so-called \emph{back-translation} of monolingual data \citep{sennrich-etal-2016-improving} is undoubtedly the most prevalent one. This approach uses a target-to-source MT system to translate monolingual data in the target language into the source language.  The generated synthetic parallel data can be used together with the original parallel data to increase the size of the training data, and eventually to obtain better NMT systems.

Nonetheless, generating synthetic parallel data in large quantity with this approach also requires a large quantity of monolingual data. For most domains in most languages, however, a large quantity of monolingual data is unavailable and thus generating large synthetic parallel data through back-translation is impracticable.

In this preliminary work, we propose a new approach that leverages small in-domain monolingual data to generate large synthetic in-domain parallel data. We demonstrate that synthetic in-domain monolingual data generated by a GPT-2 model  \citep{Radford2019LanguageMA}, fine-tuned on our very small in-domain monolingual data, can be successfully translated by NMT to generate synthetic in-domain parallel data. Our results on three language pairs and five domains show improvements in BLEU for all configurations when using our synthetic data to train NMT. We also show that this approach can be used in extreme adaptation for personalized NMT.

\begin{figure*}[t]
	\small
	\begin{flushleft}
	(a) Generated by GPT-2 fine-tuned on Medical data
	\end{flushleft}
	\begin{displayquote}\em
	$\ldots$	Because of methodological differences we could not obtain a comparable result for 17S viral nucleic acids or 16S viral nucleic acid using different methods.\\
		The SARI statement: A measure of the quality of health services, including the availability of drugs, is a basic criterion for measuring the quality of health services system.\\
		The 12 patients seen at the DCP + IC applied for six of these six HDCP methods (75\%) successfully completed their pharmacy duties as per the guidelines.$\ldots$
	\end{displayquote}	

		\begin{flushleft}
		(b) Generated by GPT-2 fine-tuned on IT data
	\end{flushleft}
	\begin{displayquote}\em
$\ldots$You can use the Page Colors application that you can find on Google+\\
The maps of Portugal are free but you can acquire maps for other countries\\
You can use the program Ringtone Maker which you can find on Google$\ldots$
	\end{displayquote}	
		\begin{flushleft}
		(c) Generated by GPT-2 fine-tuned on tweets (natural disaster domain)
\end{flushleft}
\begin{displayquote}\em
$\ldots$A volcanic eruption in 1815 set off a massive effluence that sowed \#wildfire on the west coast!Thanks NSW \#NSWgovernors for treating these \#shills \\
4.4 earthquake occurred near Negros Region, Chile at 22: 10 UTC! \#earthquake \#NegrosRegion\\
Day: Malta - Black cloud surrounded by rain. 16: 35 JST 14 / 16 / 19 - 21: 45 JST 15 / 16 / 19 - 17: 00 JST$\ldots$
\end{displayquote}	

		\begin{flushleft}
		(d) Generated by GPT-2 not fine-tuned
\end{flushleft}
\begin{displayquote}\em
$\ldots$On Thursday, fossil fuels minister Emmanuel Ponting said the year 2000 was the year that the entire human genome was analyzed and "explored for new genes."\\
Consider the mercenary work that Columbia University puts in.\\
Such coins have been suggested by Buzzfeed tech reporter Alex Seitz, who wrote a very thorough investigation into the issue.$\ldots$
\end{displayquote}	
\caption{\label{fig:example} Examples of three raw consecutive lines inside one sequence generated by different GPT-2 models. We manually added ``$\ldots$'' to show the reader that they are extracts from a sequence. GPT-2 models, fine-tuning data, and hyper-parameters used to obtain these examples are presented in Section \ref{sec:exp}.}
\end{figure*}

\section{Motivation}
This work relies on three assumptions:
\begin{itemize}
    \item GPT models generates mostly correct sentences.
    \item Sentences generated by a GPT model exhibit some of the characteristics of the in-domain data on which the model has been fine-tuned, even if the data is small.
    \item NMT training is robust, to some extent, to the noise in the texts generated by GPT models.
\end{itemize}

For our two first assumptions, we can obtain some hints on their validity by manually checking sentences generated by fine-tuned GPT-2 models. Examples of such sentences are presented in Figure~\ref{fig:example}. We can see with these examples that GPT-2 models successfully generate sentences that are mostly correct and present characteristics of the domain on which they have been fine-tuned.

For our third assumption, we rely on previous work that shows that back-translations in which artificial noise has been injected can improve translation quality when used for training NMT \citep{edunov-etal-2018-understanding}.

\section{Synthesizing Large Parallel Data Leveraging Small Monolingual Data}
\subsection{Requirements}
Our method has few requirements in terms of data that make it applicable in most MT scenarios. Precisely, we need the following three types of data:
\begin{itemize}
    \item a GPT model or large (general-domain) monolingual data: in this preliminary work, we only exploit the smallest GPT-2 model released by OpenAI. For the future, we plan to experiment with in-house GPT models, trained on large general-domain monolingual data.
    \item small in-domain monolingual data: most of our experiments use 50k sentences for each target domain, but experiments in extreme adaptation for personalized NMT shows that our method is useful even when only hundreds of sentences are available.
    \item some parallel data: all our experiments use at least 156k sentence pairs.
\end{itemize}
\subsection{Synthetic Monolingual Data}
We use GPT-2 \citep{Radford2019LanguageMA}\footnote{\url{https://github.com/openai/gpt-2}} to generate synthetic monolingual data. GPT models are auto-regressive Transformer \citep{NIPS2017_7181} decoders. Given some context, or no context at all if this is the first token of the sequence, the model predicts the next token. 

To generate texts in a particular domain, we fine-tuned a given GPT-2 model on a small amount of texts in the target domain and language. Since GPT-2 is efficient for text generation, we can generate millions of in-domain monolingual sentences.


\subsection{Synthetic Parallel Data}
Once the synthetic monolingual data are generated, it can be used in NMT as any other monolingual data. In this work, we demonstrate its usefulness through back-translation \citep{sennrich-etal-2016-improving} and forward translation to generate in-domain synthetic parallel data. 

For back-translation, we adopted the tagged approach \citep{caswell-etal-2019-tagged} that has been shown to provide better results, especially for translating texts that are not translationese \citep{marie-etal-2020-tagged}. In this configuration, the target side of the synthetic parallel data was generated by GPT-2, in English, and the source side by NMT.

For forward translation, we did not use tags. In this configuration the source side was generated by GPT-2, in English, while the target side was obtained through NMT.
Forward translation is known to underperform back-translation \citep{bogoychev2019domain}. Nonetheless, since we do not have GPT-2 models in other languages than English, we could only exploit synthetic monolingual data for translation directions with English on the source side through forward translation.

\section{Experiments}
\label{sec:exp}
\subsection{Data}
\subsubsection{Training}
We trained NMT systems for English--German (En-De), English--French (En-Fr), and Japanese--English (Ja-En) on the following parallel data (numbers of sentence pairs are given after pre-processing described in Section \ref{sec:prepro}):
\begin{itemize}
	\item En-De: WMT17\footnote{\url{http://statmt.org/wmt17/translation-task.html}} parallel data (5.1M sentence pairs)
	\item En-Fr: WMT14\footnote{\url{http://statmt.org/wmt14/translation-task.html}} (32.7M  sentence pairs)
	\item En-Ja: Training parallel data provided in the MTNT dataset \citep{michel-neubig-2018-mtnt}\footnote{\url{http://www.cs.cmu.edu/~pmichel1/mtnt/}} which is a concatenation of three different datasets (TED talks\footnote{\url{https://wit3.fbk.eu/}}, The Kyoto Free Translation Task (KFTT)\footnote{\url{http://www.phontron.com/kftt/}}, and JESC\footnote{\url{https://nlp.stanford.edu/projects/jesc/}}) (3.9M sentence pairs)
\end{itemize}

\subsubsection{Validation}
We used one validation dataset, for each language pair, to select the best model after training NMT (see Section \ref{sec:prepro}):

\begin{itemize}
	\item En-De: WMT16 newstest\footnote{\url{http://data.statmt.org/wmt20/translation-task/dev.tgz}} (2,999 sentence pairs)
	\item En-Fr: WMT13 newstest\footnote{\url{http://data.statmt.org/wmt20/translation-task/dev.tgz}} (3,000 sentence pairs)
	\item En-Ja: Validation data provided in the MTNT dataset\footnote{\url{http://www.cs.cmu.edu/~pmichel1/mtnt/}} that is a concatenation of data from TED Talks, KFTT, and JESC corpora (4,451 sentence pairs)
\end{itemize}

\subsubsection{Test}
We used several datasets from different domains for evaluating the translation quality of our NMT systems  for each language pair:
\begin{itemize}
	\item En-De:
	\begin{itemize}
		\item News domain: WMT17 news translation task\footnote{\url{http://data.statmt.org/wmt20/translation-task/dev.tgz}} (3,004 sentence pairs)
		\item Medical domain: WMT14 medical translation task, khresmoi summary\footnote{\url{http://www.statmt.org/wmt14/medical-task/khresmoi-summary-test-set.tgz}} (1,000 sentence pairs)
		\item IT domain: WMT16 IT translation task, batch 3\footnote{\url{http://data.statmt.org/wmt16/it-translation-task/wmt16-it-task-references.tgz}} (1,000 sentence pairs)
	\end{itemize}
	\item En-Fr:
\begin{itemize}
	\item News domain: WMT14 news translation  task\footnote{\url{http://data.statmt.org/wmt20/translation-task/dev.tgz}} (3,003 sentence pairs)
	\item Medical domain: WMT14 medical translation task, khresmoi summary\footnote{\url{http://www.statmt.org/wmt14/medical-task/khresmoi-summary-test-set.tgz}} (1,000 sentence pairs)
	\item Reddit domain: MTNT test sets,\footnote{\url{http://www.cs.cmu.edu/~pmichel1/mtnt/}} one for each translation direction (for En$\rightarrow$Fr: 1,020 sentence pairs, for Fr$\rightarrow$En: 1,022 sentence pairs)
\end{itemize}
	\item En-Ja:
\begin{itemize}
	\item News domain: ALT test set\footnote{\url{https://www2.nict.go.jp/astrec-att/member/mutiyama/ALT/}} (1,018 sentence pairs)
	\item Reddit domain: MTNT test sets,\footnote{\url{http://www.cs.cmu.edu/~pmichel1/mtnt/}} one for each translation direction (for En$\rightarrow$Ja: 1,002 sentence pairs, for Ja$\rightarrow$En: 1,001 sentence pairs)
	\item Twitter natural disaster domain: Tweets test set compiled and translated by ourselves (not publicly available) (1,400 sentence pairs)
\end{itemize}
\end{itemize}

\subsubsection{English Monolingual Data}
\label{sec:monodata}
English monolingual data are used as a source for back/forward translation and for fine-tuning GPT-2. There is one dataset for each domain:
\begin{itemize}
	\item News domain: News Crawl 2019\footnote{\url{http://data.statmt.org/news-crawl/en/news.2019.en.shuffled.deduped.gz}} (1M lines for backward/forward translation, 50k lines for GPT-2 fine-tuning)
	\item IT domain: English side of the training parallel data provided for the WMT16 IT translation task, batch 1 and batch 2\footnote{\url{http://ufallab.ms.mff.cuni.cz/~popel/batch1and2.zip}}, (2k lines for backward/forward translation and GPT-2 fine-tuning)
	\item Medical domain: English side of the \mbox{En-Fr} EMEA parallel data\footnote{\url{http://opus.lingfil.uu.se/download.php?f=EMEA/en-fr.txt.zip}} provided for the WMT14 medical translation task (100k lines for backward/forward translation, 50k lines for GPT-2 fine-tuning)
	\item Reddit domain: English data crawled with the Reddit API (1M lines for backward/forward translation, 50k lines for GPT-2 fine-tuning)
	\item Twitter natural disaster domain: English tweets crawled with the Twitter API with the same keywords used to crawled the English tweets of the test set (not publicly released) (148k lines for backward/forward translation, 50k lines for GPT-2 fine-tuning)
\end{itemize}
\subsection{Framework and Settings}
\label{sec:prepro}
We exploited GPT-2 through the gpt-2-simple framework.\footnote{\url{https://github.com/minimaxir/gpt-2-simple}}
We did not perform any pre-processing on the monolingual data used for fine-tuning GPT-2.  For NMT, we tokenized and truecased all the data in English, French, and German, with the Moses toolkit \citep{koehn-etal-2007-moses}.\footnote{\url{https://github.com/moses-smt/mosesdecoder}} The truecaser has been trained on 1M lines randomly sampled from the News Crawl corpora in each language.

For NMT, training data, validation data, and source side test set are all segmented into subword units. We used byte-pair encoding (BPE) \citep{sennrich-etal-2016-neural}\footnote{\url{https://github.com/rsennrich/subword-nmt}} for English, German, and French, separately trained on 10M lines from the News Crawl 2019 corpora\footnote{\url{http://data.statmt.org/news-crawl/}} for each language to learn 32k BPE operations. For Japanese, we used SentencePiece \citep{kudo-richardson-2018-sentencepiece}\footnote{\url{https://github.com/google/sentencepiece}} to learn 16k sentence pieces also from the News Crawl 2019 corpus.

We used Marian \citep{junczys-dowmunt-etal-2018-marian}\footnote{\url{https://marian-nmt.github.io/}, version v1.7.6 1d4ba73 2019-05-11 17:16:31 +0100} for NMT with standard hyper-parameters for training (see Table~\ref{tab:marianparam}).

\begin{table}[t]
\centering
\small
\begin{tabular}{p{7cm}}
\toprule
\texttt{--type transformer --max-length 120 --mini-batch-fit   --valid-freq 5000 --save-freq 5000 --workspace 10000 --disp-freq 500  --beam-size 12 --normalize=1 --valid-mini-batch 16 --overwrite  --early-stopping 5  --cost-type=ce-mean-words  --valid-metrics ce-mean-words bleu --keep-best   --enc-depth 6 --dec-depth 6   --transformer-dropout 0.1  --learn-rate 0.0003   --lr-warmup 16000   --lr-decay-inv-sqrt 16000 --lr-report   --label-smoothing 0.1  --devices 0 1 2 3 4 5 6 7  --optimizer-params 0.9 0.98 1e-09 --clip-norm 5  --sync-sgd  --exponential-smoothing}\\
\bottomrule
\end{tabular}
\caption{\label{tab:marianparam} Hyper-parameters of Marian used for training our NMT systems.}
\end{table}
We performed decoding with a beam size of 12 and a length normalization at 1.0.

For evaluation, we used SacreBLEU \citep{post-2018-call}\footnote{\url{https://github.com/mjpost/sacrebleu}} and report on BLEU scores \cite{papineni-etal-2002-bleu} for English, French, and German, and chrF scores \cite{popovic-2015-chrf} for Japanese. Before evaluation, we post-processed the NMT output by undoing BPE and SentencePiece subword segmentations. Then, except for Japanese, we detokenized and detruecased the output with Moses.

\subsection{Results with Back-translation}
\label{sec:exptbt}
The performance of our NMT systems trained on several different sets of back-translations is shown in Table \ref{tab:result_tbt}.

First, we assessed to what extent the human-made in-domain monolingual data used for fine-tuning GPT-2 are useful for back-translation. As we can see, despite the small size of the data, it improves BLEU compared to the baseline systems for all configurations. When using all the human-made in-domain monolingual data, or up to 1M sentences, BLEU improvements are even larger for almost all configurations (except for Ja$\rightarrow$En, Reddit). This result confirms the usefulness of exploiting more in-domain monolingual data through back-translation when available.

Using 1M sentences generated by a GPT-2 model that is not fine-tuned leads to lower BLEU scores than using all the human-made in-domain monolingual data (except for Ja$\rightarrow$En, Reddit).

The two last rows give the results of our approach: they use human-made in-domain monolingual data only up to 50k sentences for fine-tuning the GPT-2 model, but millions of synthetic monolingual data.  They show that the back-translations of the monolingual data generated by the fine-tuned GPT-2 model are useful. We obtained better, or comparable, BLEU scores when using the back-translations of our synthetic monolingual data to train NMT systems than when using the back-translations of human-made monolingual data. Using more synthetic monolingual data (last row) also tends to lead to better BLEU scores (except for Ja$\rightarrow$En, Reddit and Twitter).

\begin{table*}[t]
    \centering
    \small
    \begin{tabular}{l@{~}cccccccccc}
    \toprule
    \multirow{2}{*}{System} &Back-translated  & \multicolumn{3}{c}{De$\rightarrow$En} &  \multicolumn{3}{c}{Fr$\rightarrow$En} &  \multicolumn{3}{c}{Ja$\rightarrow$En} 
          \\
       & Data &  News & Medical & IT & News & Medical & Reddit  & News & Reddit & Twitter \\ 
    \midrule   
    Baseline & none & 32.9&36.4&42.0&36.6&48.4&34.5&14.5&7.8&5.5
 \\
    + H-TBT &fine-tuning &34.2 &40.2&42.7&37.1&48.9&34.7&17.2&8.3&16.6
 \\
    + H-TBT & all  & 35.8&40.7&43.4&37.4&49.6&35.9&22.1&8.0&17.1
 \\
    \midrule
    + GPT\_notft-TBT & 1M sentences & 34.6&37.3&41.9&37.1&48.5&34.7&20.0&8.6&9.8
 \\
    + GPT-TBT & 1M sentences & 35.5&42.6 & 42.6&37.4&49.3 &35.7&20.9&9.3&17.7
\\
    + GPT-TBT & 10M sentences & 35.5&42.9&44.6&37.8&50.3&36.9&22.3&8.7&15.9
\\

\bottomrule

    \end{tabular}
    \caption{BLEU scores of our NMT systems translating into English, for each domain. ``H-TBT'' denotes systems trained on the back-translated human-made monolingual data (the data used for fine-tuning GPT or all the monolingual data described in Section \ref{sec:monodata}). ``GPT-TBT'' denotes systems trained on the back-translation of either 1M or 10M monolingual sentences generated by a GPT-2 model fine-tuned on the in-domain monolingual data. ``GPT\_notft-TBT'' denotes a configuration in which GPT-2 has not been fine-tuned.}
    \label{tab:result_tbt}
\end{table*}
\subsection{Results with Forward Translation}
\label{sec:expft}
We performed similar experiments as in Section \ref{sec:exptbt} but with forward translation instead of back-translation. Our results are shown in Table \ref{tab:result_ft}.

We did not observe consistent improvements of BLEU and chrF scores when exploiting human-made monolingual data (H-FT configurations). Increasing the amount of monolingual data can also decrease or increase BLEU and chrF scores. Our approach (GPT-FT) leads to better, or similar, scores than the H-FT configurations that use all the human-made monolingual data.

We conclude that forward translations perform reasonably well, but not consistently, in these configurations and that GPT-2 models in other languages than English would be necessary to properly evaluate to which extent our approach can improve BLEU and chrF scores when English is not the target language.

\begin{table*}[t]
    \centering
    \small
    \begin{tabular}{l@{~}cccccccccc}
    \toprule
    \multirow{2}{*}{System} &Translated  & \multicolumn{3}{c}{En$\rightarrow$De (BLEU)} &  \multicolumn{3}{c}{En$\rightarrow$Fr (BLEU)} &  \multicolumn{3}{c}{En$\rightarrow$Ja (chrF)} 
          \\
       & Data &  News & Medical & IT & News & Medical & Reddit  & News & Reddit & Twitter \\ 
    \midrule   
    Baseline & none & 27.3&28.8&37.4&36.3&40.9&25.5&0.2436&0.1419&0.0987
 \\
    + H-FT &fine-tuning & 27.9&29.6&38.6&36.5&40.9&23.3&0.2643&0.1400&0.0839
\\
    + H-FT & all  & 27.9&29.7&38.6&36.2&41.6&23.4&0.2847&0.1348&0.0845
\\
    \midrule
    + GPT\_notft-FT & 1M sentences &27.4 &28.7 &36.7 &36.0 &40.5 &22.5 &0.2479 &0.1301 &0.0799
\\
    + GPT-FT & 1M sentences & 27.9&29.6&39.1&36.2&42.0&23.1&0.2513&0.1324&0.0832
\\
    + GPT-FT & 10M sentences & 28.0&30.1&38.9&36.3&42.3&23.3&0.2749&0.1321&0.0810
 \\

\bottomrule

    \end{tabular}
    \caption{BLEU and chrF scores of our NMT systems translating from English, for each domain. ``H-FT'' denotes systems trained on forward-translated human-made monolingual data (the data used for fine-tuning GPT or all the monolingual data described in Section \ref{sec:monodata}). ``GPT-FT'' denotes systems trained on the forward-translation of either 1M or 10M monolingual sentences generated by a GPT-2 model fine-tuned on the in-domain monolingual data. ``GPT\_notft-FT'' denotes a configuration in which GPT-2 has not been fine-tuned.}
    \label{tab:result_ft}
\end{table*}

\section{Extreme Adaptation for Personalized NMT}
The objective of extreme adaptation for personalized NMT is to adapt a given NMT system so that it can better translate texts written or spoken by a specific person. Ideally, for such a task, we would require an amount as large as possible of parallel data in which one side are texts written or spoken by the target person in order to personalize our NMT system. Obviously, such a large data does not exist and would be too costly to create. Thus, we propose to synthesize such a data with our approach. The main difference with the domain adaptation scenarios presented in Section \ref{sec:exp} is that we cannot even expect to obtain thousands of sentences of texts written by the target person to fine-tune GPT-2.

For our extremely personalized NMT experiments, we used the Speaker Annotated TED Talks (SATED) corpus \citep{michel-neubig-2018-extreme}\footnote{\url{http://www.cs.cmu.edu/~pmichel1/sated/}} available for:
\begin{itemize}
	\item English--German (En-De): 156k sentence pairs, 1,670 speakers
	\item English--Spanish (En-Es): 183k sentence pairs, 1,922 speakers
	\item English--French (En-Fr): 178k sentence pairs, 1,887 speakers
\end{itemize}

Each sentence pair is provided with a tag that identifies the speaker. Note that this corpus is already pre-processed: tokenized and lower-cased.
Validation data and test data contain two sentence pairs per speaker.
In order to generate synthetic monolingual data for each specific speaker, we exploit the speaker tag by concatenating it to the English side of the parallel data and then use the resulting data to fine-tune the GPT-2 model. Through fine-tuning, we assume that GPT-2 learns the characteristics of each individual speaker by relying on the speaker tag.
At decoding time, we then expect GPT-2 to generate texts for a particular speaker when prompting it with its speaker tag.

\begin{table*}
    \centering
    \small
    \begin{tabular}{lcccc}
    \toprule
       System & De$\rightarrow$En &  Es$\rightarrow$En  & Fr$\rightarrow$En  \\
         \midrule 
         Baseline & 24.6 & 32.2 & 29.5 \\
          Speaker Tags& 24.7 & 32.2 & 29.9 \\
          \midrule
          Speaker Tags + GPT-TBT  & 27.6 & 34.6 & 32.2\\ 
          Speaker Tags + GPT\_speaker-TBT & 28.5 & 35.6 & 32.4 \\ 
    \bottomrule     
    \end{tabular}
    \caption{BLEU scores of our NMT systems for the SATED translation task. ``Speaker Tags'' denotes the use of the speaker tags to tag each sentence pair in the given parallel data and each source sentence in the test data. ``GPT-TBT'' denotes systems trained on back-translation of 500k sentences generated by a GPT-2 model fine-tuned on the English side of the SATED parallel data. `GPT\_speaker-TBT'' is similar to `GPT-TBT,'' except (1) that the data used for fine-tuning the GPT-2 model are also tagged with the speaker tag and (2) that each sentence generated by the fine-tuned GPT-2 model is tagged with a speaker tag.}
    \label{tab:epnmt}
\end{table*}

The results of our experiments are presented in Table \ref{tab:epnmt}.
In addition to the a vanilla baseline NMT system, we used one of the adaptation approach (second row) that uses the parallel data with the speaker tag concatenated to the source sentence to train NMT systems \citep{michel-neubig-2018-extreme}. BLEU scores with this approach are close to the scores of the baseline system.
Then, we tried our approach similarly to our experiments in Section \ref{sec:exptbt} (third row). We fine-tuned GPT-2 on the English side of the SATED parallel data and generated 500k sentences with the fine-tuned model. Then, we back-translated the generated data with En$\rightarrow\ast$ baseline systems to obtain synthetic parallel data for the three language pairs and concatenated it to the speaker tagged SATED parallel data exploited in our experiments of the second row. The NMT systems trained on the resulting data improve the BLEU scores by several BLEU points for all the translation directions.
In the last row, we finally report on the results exploiting the speaker tags also when fine-tuning the GPT-2 model. We generated 500k sentences, randomly prompting GPT-2 with one of the speaker tags, and exploited the resulting speaker-tagged monolingual data as for the other models. BLEU scores are further improved, implying that GPT-2 successfully exploits the speaker tag to generate better synthetic data for each speaker.

\section{Conclusion and Future Work}
In this preliminary work, we showed that our approach can leverage small in-domain monolingual data produced by human to generate a large synthetic in-domain parallel data. Even though the synthetic parallel data are entirely synthetic, as opposed to a standard backward/forward translation, we obtained improvements in BLEU scores in all our configurations when using the generated data to train NMT systems. We also reported on successful experiments in extreme adaptation for personalized NMT.

In our future work, we would like to perform an in-depth analysis to better understand our results. We will also conduct more experiments exploiting in-house GPT models for other languages.

\bibliographystyle{acl_natbib}
\bibliography{acl2020}
\end{document}